\documentclass[11pt,a4paper]{article}
\usepackage[hyperref]{emnlp-ijcnlp-2019}
\usepackage{times}
\usepackage{latexsym}
\usepackage{graphicx}
\usepackage{amsfonts,amssymb}
\usepackage{booktabs}
\usepackage{amsmath}
\usepackage{subfig}
\usepackage{multirow}
\usepackage{bm}
\usepackage[marginal]{footmisc}
\usepackage{framed}
\usepackage{color}
\usepackage{colortbl}
\DeclareMathOperator*{\argmax}{argmax}

\DeclareMathOperator*{\concat}{concat}
\DeclareMathOperator*{\attention}{attention}
\DeclareMathOperator*{\softmax}{softmax}
\DeclareMathOperator*{\head}{head}

\usepackage[toc,page]{appendix}
\usepackage{url}

\title{Hierarchically-Refined Label Attention Network for Sequence Labeling}

\author{
  Leyang Cui$^{1,2,3}$ \and Yue Zhang$^{2,3}$ \\
  $^1$Zhejiang University \\
  $^2$School of Engineering, Westlake University \\
  $^3$Institute of Advanced Technology, Westlake Institute for Advanced Study \\
  {\tt cuileyang@westlake.edu.cn, yue.zhang@wias.org.cn}
  }

\date{}

\begin{document}

\aclfinalcopy
\maketitle

\begin{abstract}
CRF has been used as a powerful model for statistical sequence labeling. For neural sequence labeling, however, BiLSTM-CRF does not always lead to better results compared with BiLSTM-softmax local classification. This can be because the simple Markov label transition model of CRF does not give much information gain over strong neural encoding. For better representing label sequences, we investigate a hierarchically-refined label attention network, which explicitly leverages label embeddings and captures potential long-term label dependency by giving each word incrementally refined label distributions with hierarchical attention. Results on POS tagging, NER and CCG supertagging show that the proposed model not only improves the overall tagging accuracy with similar number of parameters, but also significantly speeds up the training and testing compared to BiLSTM-CRF.
\end{abstract}
\section{Introduction}

Conditional random fields (CRF) \cite{Lafferty:2001:CRF:645530.655813} is a state-of-the-art model for statistical sequence labeling \cite{Toutanova:2003:FPT:1073445.1073478,Peng:2004:CSN:1220355.1220436,Ratinov:2009:DCM:1596374.1596399}.
Recently, CRF has been integrated with neural encoders as an output layer to capture label transition patterns \cite{P15-1109,P16-1101}.
This, however, sees mixed results.
For example,  previous work \cite{reimers2017optimal,yang2018design} has shown that BiLSTM-softmax gives better accuracies compared to BiLSTM-CRF for part-of-speech (POS) tagging. In addition, the state-of-the-art  neural Combinatory Categorial Grammar (CCG) supertaggers do not use CRF \cite{P15-2041,N16-1026}.

One possible reason is that the strong representation power of neural sentence encoders such as BiLSTMs allow models to capture implicit long-range label dependencies from input word sequences alone \cite{DBLP:journals/tacl/KiperwasserG16,dozat2016deep,C18-1011}, thereby allowing the output layer to make local predictions. 
In contrast, though explicitly capturing output label dependencies, CRF can be limited by its Markov assumptions, particularly when being used on top of neural encoders. In addition, CRF can be computationally expensive when the number of labels is large, due to the use of Viterbi decoding.

One interesting research question is whether there is neural alternative to CRF for neural sequence labeling, which is both faster and more accurate. To this question, we investigate a neural network model for output label sequences. In particular, we represent each possible label using an embedding vector, and aim to encode sequences of label distributions using a recurrent neural network. One main challenge, however, is that the number of possible label sequences is exponential to the size of input. This makes our task essentially to represent a full-exponential search space without making Markov assumptions.

\begin{figure}[!t]
\setlength{\abovecaptionskip}{3pt}
\setlength{\belowcaptionskip}{-3pt}
\centering
\includegraphics[height=3.5cm, width=7.5cm]{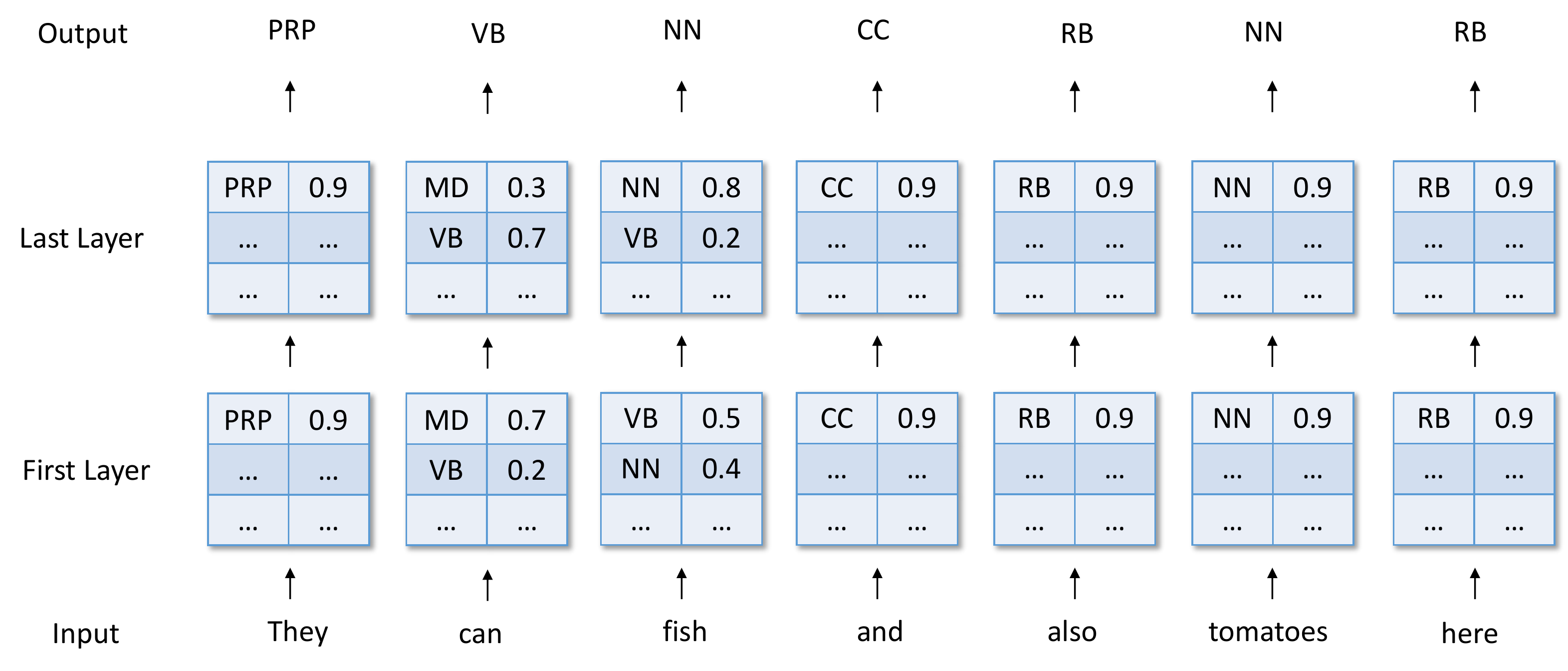}
\caption{Visualization of hierarchically-refined Label Attention Network for POS tagging. The numbers indicate the label probability distribution for each word.}
\label{demo}
\end{figure}

We tackle this challenge using a hierarchically-refined representation of marginal label distributions. As shown in Figure~\ref{demo}, our model consists of a multi-layer neural network. In each layer, each input words is represented together with its marginal label probabilities, and a sequence neural network is employed to model unbounded dependencies. The marginal distributions space are refined hierarchically bottom-up, where a higher layer learns a more informed label sequence distribution based on information from a lower layer.

For instance, given a sentence ``They$_1$ can$_2$ fish$_3$ and$_4$ also$_5$ tomatoes$_6$ here$_7$'', the label distributions of the words  \textit{can}$_2$ and \textit{fish}$_3$ in the first layer of Figure~\ref{demo} have higher probabilities on the tags MD (modal verb) and VB (base form verb), respectively, though not fully confidently. The initial label distributions are then fed as the inputs to the next layer, so that long-range label dependencies can be considered. In the second layer, the network can learn to assign a noun tag to \textit{fish$_3$} by taking into account the highly confident tagging information of \textit{tomatoes$_6$ (NN)}, resulting in the pattern ``\textit{can$_2$ (VB) fish$_3$ (NN)}''.

\begin{figure*}[!t]
\setlength{\abovecaptionskip}{3pt}
\setlength{\belowcaptionskip}{0pt}
\centering
\includegraphics[width=16cm]{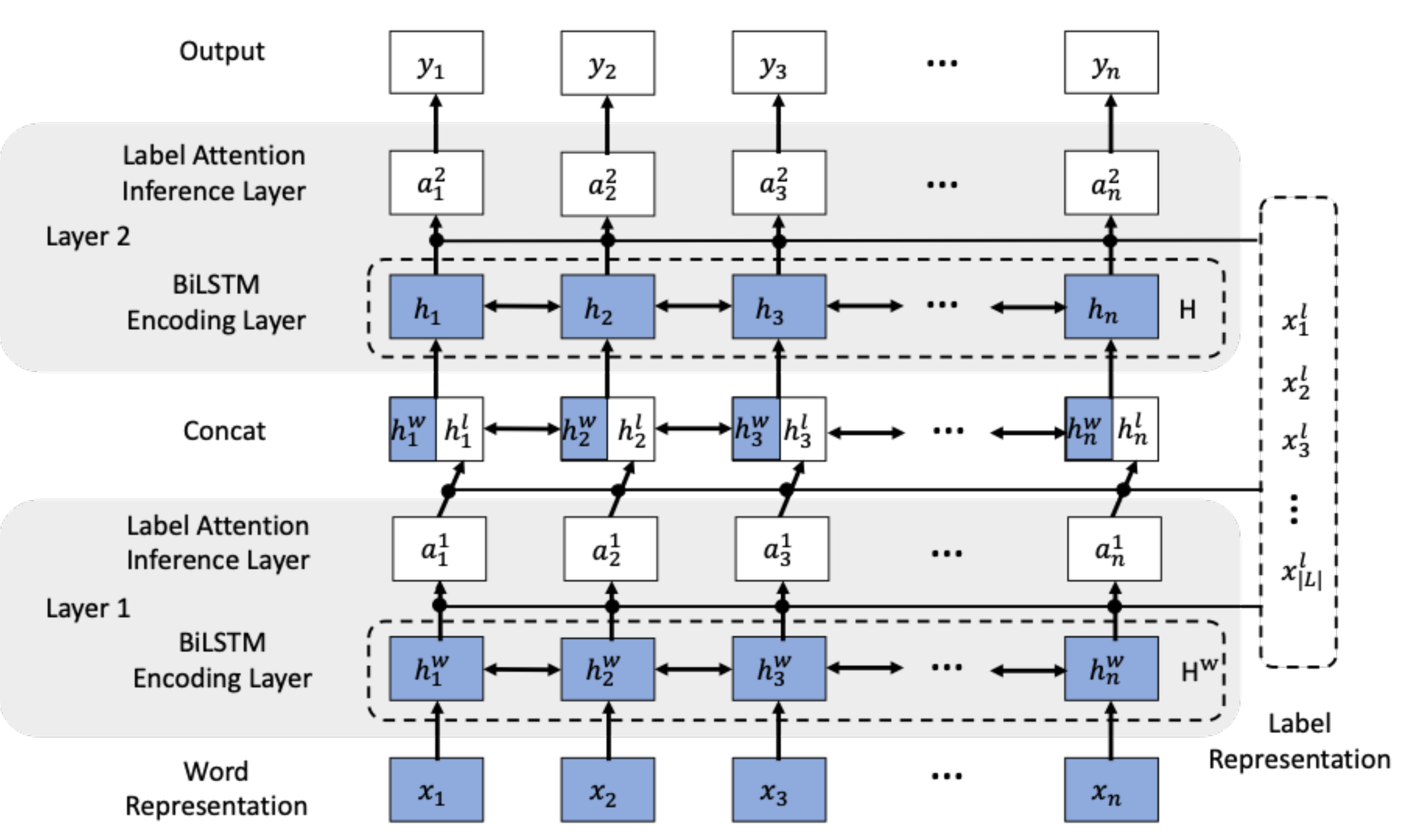}
\caption{\label{architecture} Architecture of hierarchically-refined label attention network. }
\end{figure*}

As shown in Figure~\ref{architecture}, our model consists of stacked attentive BiLSTM layers, each of which takes a sequence of vectors as input and yields a sequence of hidden state vectors together with a sequence of label distributions. The model performs attention over label embeddings \cite{N15-1030,D18-1484} for deriving a marginal label distributions, which are in turn used to calculate a weighted sum of label embeddings. Finally, the resulting packed label vector is used together with input word vectors as the hidden state vector for the current layer. Thus our model is named \textit{label attention network} (LAN). For sequence labeling, the input to the whole model is a sentence and the output is the label distributions of each word in the final layer.

BiLSTM-LAN can be viewed as a form of multi-layered BiLSTM-softmax sequence labeler. In particular, a single-layer BiLSTM-LAN is identical to a single-layer BiLSTM-softmax model, where the label embedding table serves as the softmax weights in BiLSTM-softmax, and the label attention distribution is the softmax distribution in BiLSTM-softmax. The traditional way of making a multi-layer extention to BiLSTM-softmax is to stack multiple BiLSTM encoder layers before the softmax output layer, which learns a deeper input representation. In contrast, a multi-layer BiLSTM-LAN stacks both the BiLSTM encoder layer and the softmax output layer, learning a deeper representation of both the input and candidate output sequences.

On standard benchmarks for POS tagging, NER and CCG supertagging, our model achieves significantly better accuracies and higher efficiencies than BiLSTM-CRF and BiLSTM-softmax with similar number of parameters. It gives highly competitive results compared with top-performance systems on WSJ, OntoNotes 5.0 and CCGBank without external training. In addition to accuracy and efficiency, BiLSTM-LAN is also more interpretable than BiLSTM-CRF thanks to visualizable label embeddings and label distributions. Our code and models are released at \url{https://github.com/Nealcly/LAN}.

\section{Related Work}

{\bf Neural Attention.} Attention has been shown useful in neural machine translation \cite{bahdanau2014neural}, sentiment classification \cite{D17-1047,E17-2091}, relation classification \cite{P16-2034}, read comprehension \cite{Hermann:2015:TMR:2969239.2969428}, sentence summarization \cite{D15-1044}, parsing \cite{D16-1035}, question answering \cite{C18-1215} and text understanding \cite{P16-1086}. Self-attention network (SAN) \cite{NIPS2017_7181} has been used for semantic role labeling \cite{D18-1548}, text classification \cite{P18-2123,D18-1408} and other tasks. 
Our work is similar to \citet{NIPS2017_7181} in the sense that we also build a hierarchical attentive neural network for sequence representation. The difference lies in that our main goal is to investigate the encoding of exponential \textit{label} sequences, whereas their work focuses on encoding of a \textit{word} sequence only.

{\bf Label Embeddings.} Label embedding was first used in the field of computer vision for facilitating zero-shot learning \cite{PalatucciNIPS2009,Socher:2013:ZLT:2999611.2999716,7781018}. The basic idea is to improve the performance of classifying previously unseen class instances by learning output label knowledge.
In NLP, label embeddings have been exploited for better text classification \cite{tang2015pte,nam16allintext,wang2018joint}. However, relatively little work has been done investigating label embeddings for sequence labeling. One exception is \newcite{N16-1027}, who use supertag embeddings in the output layer of a CCG supertagger, through a combination of local classification model rescored using a supertag language model. In contrast, we model deep label interactions using a dynamically refined sequence representation network. To our knowledge, we are the first to investigate a hierarchical attention network over a label space.

{\bf Neural CRF}. There has been methods that aim to speed up neural CRF \cite{tu-18}, and to solve the Markov constraint of neural CRF. In particular, \citet{ijcai2018-637} predicts a sequence of labels as a sequence to sequence problem; \citet{guo-etal-2019-star} further integrates global input information in encoding. Capturing non-local dependencies between labels, these methods, however, are slower compared with CRF. In contrast to these lines of work, our method is both asymptotically faster and empirically more accurate compared with neural CRF.
\section{Baseline}
We implement BiLSTM-CRF and BiLSTM-softmax baseline models with character-level features \cite{DosSantos:2014:LCR:3044805.3045095,N16-1030}, which consist of a word representation layer, a sequence representation layer and a local softmax or CRF inference layer.

\subsection{Word Representation Layer}
\label{sec:Word Representation Layer}
Following \citet{DosSantos:2014:LCR:3044805.3045095} and \citet{N16-1030}, we use character information for POS tagging.
Given a word sequence $w_1,w_2,...,w_n$ , where $c_{ij}$ denotes the $j$th character in the $i$th word, each $c_{ij}$ is represented using
\begin{center}
$\mathbf{x}_{ij}^c = \mathbf {e}^c(c_{ij})$
\end{center}
where $\mathbf {e}^c$ denotes a character embedding lookup table.

We adopt BiLSTM for character encoding. $x_i^c$ denotes the output of character-level encoding.

A word is represented by concatenating its word embedding and its character representation:
\begin{center}
$\mathbf{x}_{i} = [\mathbf {e}^w(w_i);\mathbf{x}_i^c]$
\end{center}
where $\mathbf e^w$ denotes a word embedding lookup table.

\subsection{Sequence Representation Layer}
\label{sec:BiLSTM}
For sequence encoding, the input is a sentence $\mathbf{x} = \{\mathbf{x}_1, \cdot \cdot \cdot ,\mathbf{x}_n\}$. Word representations are fed into a BiLSTM layer, yielding a sequence of forward hidden states $\{{\stackrel {\rightarrow}{\mathbf{h}_1^w}},\cdot \cdot \cdot,{\stackrel {\rightarrow}{\mathbf{h}_n^w}}\}$ and a sequence of backward hidden states $\{{\stackrel {\leftarrow}{\mathbf{h}_1^w}},\cdot \cdot \cdot,{\stackrel {\leftarrow}{\mathbf{h}_n^w}}\}$, respectively. Finally, the two hidden states are concatenated for a final representation
\begin{equation*}
\begin{split}
\mathbf{h}^w_i &= [\stackrel{\rightarrow}{\mathbf{h}^w_i};\stackrel{\leftarrow}{\mathbf{h}^w_i}]  \\
\mathbf{H}^w &= \{\mathbf{h}_1^w, \cdot \cdot \cdot , \mathbf{h}_n^w\}
\end{split}
\end{equation*}

\subsection{Inference Layer}
{\bf CRF.}
A CRF layer is used on top of the hidden vectors $\mathbf{H}^w$.
The conditional probabilities of label distribution sequences $y$ = $\{y_1, \cdot \cdot \cdot ,y_n\}$ is

\begin{center}
$P(y|x) = \frac{\exp(\sum_{i} (\mathbf{W}^{l_i}_{CRF} \mathbf{h_i}^w + b^{(l_{i-1},l_i)}_{CRF}))}{\sum_{{y}'}
 \exp(\sum_{i} (\mathbf{W}^{{l_i}'}_{CRF} \mathbf{h_i}^w + b^{({l}'_{i-1},{l_i}')}_{CRF}))}$
\end{center}

Here ${y}'$ represents an arbitrary label distribution sequence, $ \mathbf{W}^{l_i}_{CRF} $ is a model parameter specific to $l_i,$ and $b^{({l}'_{i-1},{l_i}')}_{CRF}$ is a bias specific to $l_{i-1}$ and $l_i$.

The first-order Viterbi algorithm is used to find the highest scored label sequence over an input word sequence during decoding.

{\bf Softmax.} Independent local softmax classification can give competitive result on sequence labeling \cite{reimers2017optimal}. For each node, $\mathbf{h}_i^w$ is fed to a softmax layer to find

\begin{equation}
\hat{y}_i = \softmax(\mathbf{W} \mathbf{h}_i^w + \mathbf{b})
\label{softmax}
\end{equation}
where $\hat{y}_i$ is the predicted label for $w_i$; $\mathbf{W}$ and $\mathbf{b}$ are the parameters for softmax layer.





\section{Label Attention Network}
\label{sec:Label Attention Network}



The structure of our refined label attention network is shown in Figure~\ref{architecture}. We denote our model as BiLSTM-LAN (namely BiLSTM-label attention network) for the remaining of this paper. We use the same input word representations for BiLSTM-softmax, BiLSTM-CRF and BiLSTM-LAN, as described in Section~\ref{sec:Word Representation Layer}. Compared with the baseline models, BiLSTM-LAN uses a set of BiLSTM-LAN layers for both encoding and label prediction,  instead of a set of traditional sequence encoding layers and an inference layer.

\subsection{Label Representation.}
Given the set of candidates output labels $L$ = $\{l_1, \cdot \cdot \cdot ,l_{|L|}\}$, each label $l_k$ is represented using an embedding vector:
\begin{center}
$
\mathbf{x}_k^l = \mathbf {e}^l(l_k)
$
\end{center}

\noindent where $\mathbf{e}^l$ denotes a label embedding lookup table. Label embeddings are randomly initialized and tuned during model training.

\subsection{BiLSTM-LAN Layer}
The model in Figure~\ref{architecture} consists of 2 BiLSTM-LAN layers. As discussed earlier, each BiLSTM-LAN layer is composed of a BiLSTM encoding sublayer and a label-attention inference sublayer. In paticular, the former is the same as the BiLSTM layer in the baseline model, while the latter uses multi-head attention \cite{NIPS2017_7181} to jointly encode information from the word representation subspace and the label representation subspace. For each head, we apply a dot-product attention with a scaling factor to the inference component, deriving label distributions for BiLSTM hidden states and label embeddings.

{\bf BiLSTM Encoding Sublayer.}
Denote the input to each layer as $\mathbf{x}=\{\mathbf{x}_1, \mathbf{x}_2,..., \mathbf{x}_n\}$. BiLSTM (Section \ref{sec:BiLSTM}) is used to calculate $\mathbf{H}^w \in \mathbb {R}^{n \times d_h}$, where ${n}$ and ${d_h}$ denote the word sequence length and BiLSTM hidden size (same as the dimension of label embedding), respectively.

{\bf Label-Attention Inference Sublayer.} 
For the label-attention inference sublayer, the attention mechanism produces an attention matrix $\bm{\alpha}$ consisting of a potential label distribution for each word. We define $\mathbf{Q}$ = $\mathbf{H}^w$, $\mathbf{K}$ = $\mathbf{V}$ = $\mathbf{x}^l$. $\mathbf{x}^l　\in \mathbb {R}^{|L| \times d_h}$
is the label set representation, where $|L|$ is the total number of labels. As shown in Figure~\ref{architecture}, outputs are calculated by
\begin{equation*}
\begin{split}
\mathbf{H}^l &= \attention(\mathbf{Q},\mathbf{K},\mathbf{V}) = \bm{\alpha} \mathbf{V} \\
\bm{\alpha} &= \softmax (\frac {\mathbf{QK}^T} {\sqrt{d_h}}) 
\end{split}
\end{equation*}
Instead of the standard attention mechanism above, it can be beneficial to use multi-head for capturing multiple possible of potential label distributions in parallel. 
\begin{equation*}
\begin{split}
\mathbf{H}^l &= \concat({\head},\dots,{\head}_k) + \mathbf{H}^w \\
{\head}_i &= \attention(\mathbf{QW}_i^Q,\mathbf{KW}_i^K,\mathbf{VW}_i^V) \\
\end{split}
\end{equation*}
where $\mathbf{W}_i^Q \in \mathbb {R}^{d_{h} \times \frac{d_{h}}{k}}$, $\mathbf{W}_i^K \in \mathbb {R}^{d_{h} \times \frac{d_{h}}{k}}$ and $\mathbf{W}_i^V \in \mathbb {R}^{d_{h} \times \frac{d_{h}}{k}}$ are parameters to be learned during the training, $k$ is the number of parallel heads.


The final representation for each BiLSTM-LAN layer is the concatenation of BiLSTM hidden states and attention outputs :
\begin{center}
$ \mathbf{H} = [\mathbf{H}^w;\mathbf{H}^l]$
\end{center}
$\mathbf{H}$ is then fed to a subsequent BiLSTM-LAN layer as input, if any.


{\bf Output.}
In the last layer, BiLSTM-LAN directly predicts the label of each word based on the attention weights.
\begin{center}
\begin{equation*}
\begin{array}{c}
\left[ \begin{array}{c}
\hat{y_1}^{1} \cdots \hat{y_1}^{|L|}\\
\cdot \cdots \cdot \\
\cdot \cdots \cdot \\
\cdot \cdots \cdot \\
\hat{y_i}^{1} \cdots \hat{y_i}^{|L|}
\end{array}
\right] = \bm{\alpha}
\end{array}
\end{equation*}
\end{center}
\begin{center}
$\hat{y_i} = \argmax_j (\hat{y_i}^{1}, ... ,\hat{y_i}^{n})$
\end{center}
where $\hat{y_i}^{j}$ denotes the $j$th candidate label for the $i$th word and $\hat{y_i}$ is the predicted label for $i$th word in the sequence.

\subsection{Training}
BiLSTM-LAN can be trained by standard backpropagation using the log-likelihood loss. The training object is to minimize the cross-entropy between $y_i$ and $\hat{y_i}$ for all labeled gold-standard sentences. For each sentence,
\begin{center}

$L = - \sum\limits_{i} \sum\limits_{j} y_i^j \log \hat{y}_i^j$

\end{center}
where $i$ is the word index, $j$ is the index of labels for each word.

\subsection{Complexity}
For decoding, the asymptotic time complexity is $O(|L|^2n)$ and $O(|L|n)$ for BiLSTM-CRF and BiLSTM-LAN, respectively, where $|L|$ is the total number of labels and $n$ is the sequence length. Compared with BiLSTM-CRF, BiLSTM-LAN can significantly increase the speed for sequence labeling, especially for CCG supertagging, where the sequence length can be much smaller than the total number of labels.

\subsection{BiLSTM-LAN and BiLSTM-softmax}
As mentioned in the introduction, a single-layer BiLSTM-LAN is identical to a single-layer BiLSTM-softmax sequence labeling model. In particular, the BiLSTM-softmax model is given by Eq~\ref{softmax}. In a BiLSTM-LAN model, we arrange the set of label embeddings into a matrix as follows:
\begin{equation*}
    \mathbf{x}^l = [x_1^l; x_2^l; ..., x_{|L|}^l]
\end{equation*}

A naive attention model over X has:
\begin{equation*}
    \bm{\alpha} = \softmax(\mathbf{H} \mathbf{x}^l)
\end{equation*}

It is thus straightforward to see that the label embedding matrix $\mathbf{x}^l$ corresponds to the weight matrix $\mathbf{W}$ is Eq~\ref{softmax}, and the distribution $\bm{\alpha}$ corresponds to y in Eq~\ref{softmax}.

\section{Experiments}

We empirically compare BiLSTM-CRF, BiLSTM-softmax and BiLSTM-LAN using different sequence labelling tasks, including English POS tagging, multilingual POS tagging, NER and CCG supertagging.

\begin{table}[!t]
\small
\begin{center}
\begin{tabular}{cclll}
\hline
\bf{Data}& &training & dev & test \\
\hline
& \#l & 45 & 45 & 45\\
\bf{WSJ}& \#s & 38,219 & 5,527 & 5,462\\
 & \#t & 912,344 & 131,768 & 129,654 \\
\hline
& \#l & 50 & 50 & 49\\
\bf{UD\_en} & \#s & 12,544 & 2,003 & 2,078\\
& \#t & 204,607 & 25,150 & 25,097\\

\hline
& \#l & 18 & 18 & 18\\
\bf{OntoNotes}& \#s & 59,924 & 8,528 & 8,262\\
& \#t & 1,088,503 & 147,724 & 152,728\\

\hline
& \#l & 426 & 323 & 348\\
\bf{CCGBank}& \#s & 39,604 & 1,913 & 2,407\\
& \#t & 929,552 & 45,422 & 55,371\\
\hline
\end{tabular}
\end{center}
\caption{\label{dataset} Data statistics. l:label, s:sentence, t:tokens.}
\end{table}

\begin{table}[!t]
\small
\begin{center}
\begin{tabular}{p{50pt}cccc}
\hline \bf Model & \bf {\# E$\slash$\# H} & \bf {\# L} & \bf Acc & \bf{\# Param}\\
\hline

\multirow{6}*{BiLSTM-CRF} & 200 & 1& 97.56 & 5.1M\\ 
~ &  400 & 1& 97.57 & 5.5M\\ 
~ &  400 & 2& 97.57 & 6.4M \\
~ &  400 & 3& 97.52 & 7.4M\\
~ &  600 & 2& 97.57 & 8.2M\\
~ &  600 & 3& 97.50 & 10.4M\\
\hline
\multirow{5}*{BiLSTM-LAN} & 200 & 3& 97.53 & 5.7M\\
~ &  400 & 2& 97.57 & 8.1M\\
~ &  400 & 3& 97.63 & 10.0M\\
~ &  400 & 4& 97.60 & 12.2M\\
~ &  600 & 3& 97.62 & 16.5M\\
\hline
\end{tabular}
\caption{\label{pos-dev} WSJ development set. E: label embedding size, H: hidden size, L: number of layers.
}
\end{center}
\end{table}

\begin{table}[!t]
\small
\begin{center}
\begin{tabular}{lc}
\hline \bf Model & \bf Accuracy (\%) \\
\hline
BiLSTM-CRF  &   97.57 \\
BiLSTM-softmax & 97.58 \\
BiLSTM-LAN w/o attention$^{\dag}$ & 97.59 \\
BiLSTM-LAN & 97.65 \\
\hline
\end{tabular}
\end{center}
\caption{\label{WSJ-pos-attention-lastlayer} Effect of attention layer. {\dag} denotes the model without attention sublayers except for the last layer. 
}
\end{table}

\subsection{Dataset}

For English POS tagging, we use the Wall Street Journal (WSJ) portion of the Penn Treebank (PTB) \cite{Marcus1993}, which has 45 POS tags. We adopt the same partitions used by previous work \cite{manning171189,yang2018design}, selecting sections 0-18 as the training set, sections 19-21 as the development set and sections 22-24 as the test set.
For multilingual POS tagging, we use treebanks from Universal Dependencies(UD) v2.2 \cite{silveira14gold,11234ud} with the standard splits, choosing 8 resource-rich languages in our experiments.
For NER, we use the OntoNotes 5.0 
\cite{Hovy:2006, pradhan-etal-2013-towards}. Following previous work,we adopt the official OntoNotes data split used for co-reference resolution in the CoNLL-2012 shared task \cite{pradhan-etal-2012-conll}. 
We train our CCG supertagging model on CCGBank \cite{Hockenmaier2007}. Sections 2-21 are used for training,  section 0 for development and section 23 as the test set. The performance is calculated on the 425 most frequent labels.
Table~\ref{dataset} shows the numbers of sentences, tokens and labels for training, development and test, respectively.

\subsection{Settings}

{\bf Hyper-Parameters.}
We use 100-dimensional GloVe \cite{pennington2014glove} word embeddings for English POS tagging (WSJ, UD v2.2 EN) and name entity recognition, 300-dimensional Glove word embeddings for CCG supertagging and 64-dimensional Polyglot \cite{polyglot:2013:ACL-CoNLL} word embeddings for multilingual POS tagging. The detail values of the hyper-parameters for all experiments are summarized in Appendix A.



{\bf Evaluation.}
F-1 score are used for NER. Other tasks are evaluated based on the accuracy. We repeat the same experiment five times with the same hyperparameters and report the max accuracy, average accuracy and standard deviation for POS tagging. For fair comparison, all experiments are implemented in NCRF++ \cite{yang2018ncrf} and conducted using a GeForce GTX 1080Ti with 11GB memory.

\begin{table}[!t]
\small
\begin{center}
\begin{tabular}{lcc}
\hline {\bf Model}& {\bf train (s)} & \bf test (st/s)\\
\hline
BiLSTM-CRF (POS) & 181.32 & 781.90\\
BiLSTM-LAN (POS) & 128.75 & 805.32\\
BiLSTM-CRF (CCG) & 884.67 & 599.18\\
BiLSTM-LAN (CCG) & 369.98 & 713.70\\

\hline
\end{tabular}
\end{center}
\caption{\label{pos-ccg-speed} Comparison of the training time for one iteration and decoding speed. st indicates sentences
}
\end{table}

\subsection{Development Experiments}

We report a set of WSJ development experiments to show our investigation on key configurations of BiLSTM-LAN and BiLSTM-CRF.

{\bf Label embedding size.}
Table~\ref{pos-dev} shows the effect of the label embedding size. Notable improvement can be achieved when the label embedding size increases from 200 to 400 in our model, but the accuracy does not further increase when the size increases beyond 400. We fix the label embedding size to 400 for our model.

{\bf Number of Layers.}
For BiLSTM-CRF, previous work has shown that one BiLSTM layer is the most effecitve for POS tagging \cite{P16-1101,yang2018design}. Table~\ref{pos-dev} compares different numbers of BiLSTM layers and hidden sizes. As can be seen, a multi-layer model with larger hidden sizes does not give significantly better results compared to a 1-layer model with a hidden size of 400. We thus chose the latter for the final model.

For BiLSTM-LAN, each layer learns a more abstract representation of word and label distribution sequences. As shown in Table~\ref{pos-dev}, for POS tagging, it is effective to capture label dependencies using two layers. More layers do not empirically improve the performance. We thus set the final number of layers to 3. 


\begin{table}[!t]
\small
\begin{center}
\begin{tabular}{lcc}
\hline \bf Model & \bf Mean$\pm$std &\bf Max \\
\hline
BiLSTM-CRF${}^{\dag}$  &  97.47$\pm$0.02 & 97.49 \\
BiLSTM-CRF${}^{\ddag}$  &  97.50$\pm$0.03 & 97.51 \\
BiLSTM-softmax${}^{\dag}$  &  97.48$\pm$0.02 & 97.51 \\
BiLSTM-LAN & \bf{97.58}$\pm$0.04 & \bf{97.65} \\
\hline
\end{tabular}
\end{center}
\caption{\label{pos-result} Result for POS tagging on WSJ. $\dag$ \citet{yang2018design} and $\ddag$ \citet{N18-1089} are baseline models re-implemented in NCRF++ \cite{yang2018ncrf}. Our results are same as Table 6 of \citet{yang2018design}.
}
\end{table}

\begin{table}[!t]
\small
\begin{center}
\begin{tabular}{ll}
\hline \bf Model & \bf Accuracy \\
\hline
\citet{P16-2067} &  97.22 \\
\citet{journals/corr/HuangXY15} & 97.55 \\
\citet{P16-1101} & 97.55 \\
\citet{Liu2017} & 97.53 \\
\citet{yang2018design} & 97.51 \\
\citet{P18-1030} & 97.55 \\
\citet{N18-1089} & 97.58 \\
\citet{xin-etal-2018-learning} & 97.58 \\
\hline
Transformer-softmax \cite{guo-etal-2019-star} & 97.04 \\
BiLSTM-softmax \cite{yang2018design} & 97.51 \\
BiLSTM-CRF \cite{yang2018design} & 97.51 \\
BiLSTM-LAN & \bf{97.65} \\
\hline
\end{tabular}
\end{center}
\caption{\label{pos-top} Main results on WSJ.}
\end{table}

\begin{figure}[t]
\setlength{\abovecaptionskip}{3pt}
\setlength{\belowcaptionskip}{-3pt}
\centering
\includegraphics[width=8cm]{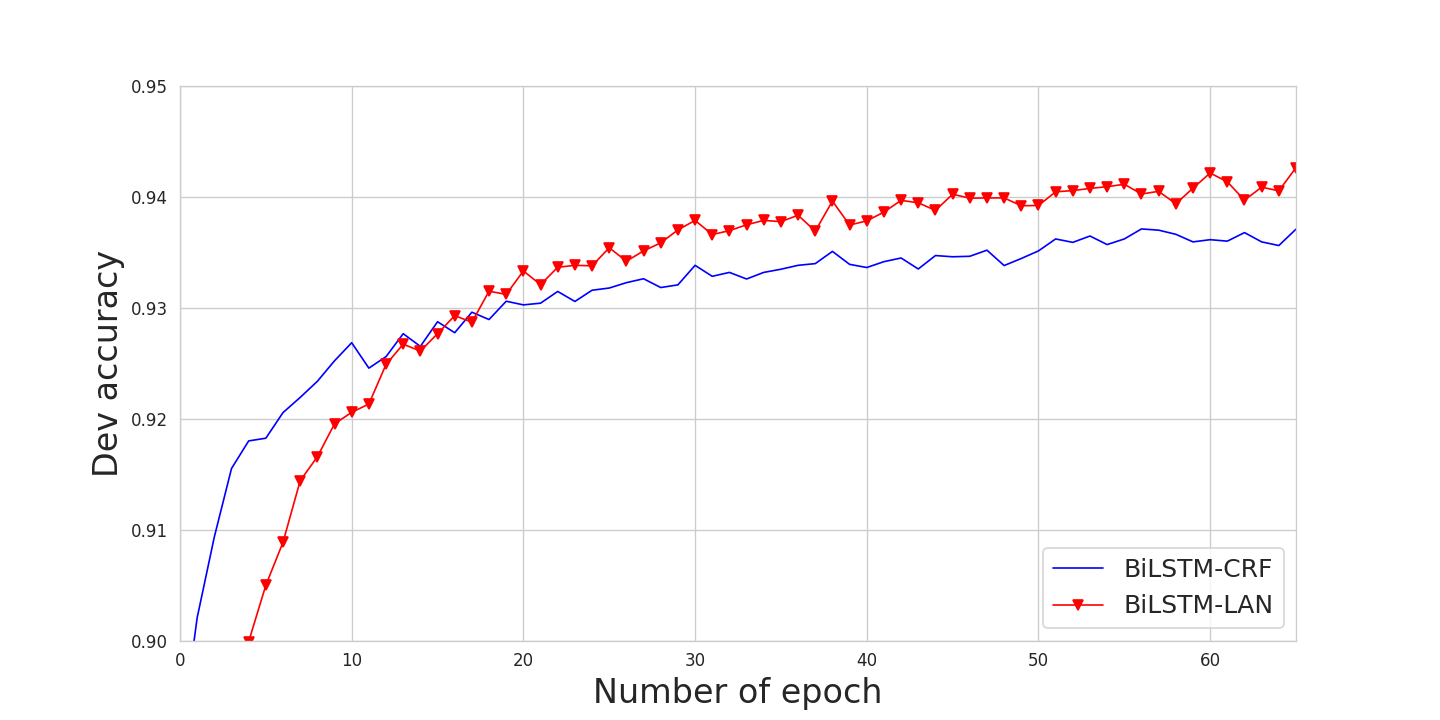}
\caption{\label{dev_accuracy_curve} Training on the WSJ development set.}
\end{figure}

\begin{table*}[!t]
\small
\begin{center}
\begin{tabular}{llcccccccc}
\hline
&& cs & da & en & fr & nl & no & pt & sv \\
\hline
\multirow{2}*{BiLSTM-CRF} & mean & 98.42 & 95.77 & 95.41 & 96.94 & 94.65 & 97.07 & 97.78 & 96.06 \\
~& $\pm$ std & \, 0.03 & \, 0.12 & \, 0.06 & \, 0.08 & \, 0.11 & \, 0.11 & \, 0.04 & \, 0.07\\
\cite{N18-1089} &training(s)& 268.74 \, & 18.17 & 58.20 & 70.10 & 44.49 & 56.06 & 51.59 & 15.54\\
\hline
\multirow{3}*{BiLSTM-softmax} & mean & 98.48 & 95.90 & 95.36 & 97.01 & 94.76 & 97.26 & 97.78 & 95.98
\\
~& $\pm$ std & \, 0.04 & \, 0.09 & \, 0.17 & \, 0.09 & \, 0.17 & \, 0.03 & \, 0.05 & \, 0.08\\
~&training(s) & 129.14 \, & 9.27 & 25.02 & 33.65 & 23.00 & 28.84 & 23.13 & 8.16\\
\hline
\multirow{3}*{BiLSTM-LAN} & mean & \bf{98.75} & \bf{96.26} & \bf{95.59} & \bf{97.28} & \bf{94.94} & \bf{97.59} & \bf{98.04} & \bf{96.55} \\
~& $\pm$ std & \, 0.02 & \, 0.12 & \, 0.13 & \, 0.08 & \, 0.11 & \, 0.04 & \, 0.04 & \, 0.01 \\
~&training(s) & 165.64 \, & 11.32 & 33.04 & 40.48 & 29.71 & 37.06 & 27.48 & 10.40\\
\hline
\end{tabular}
\end{center}
\caption{\label{pos-wd} Multilingual POS tagging result on UD v2.2 treebanks, compared on 8 resource-rich languages.}
\end{table*}

\begin{figure*}[!t]
\centering
\subfloat[5 iterations]{\fbox{\includegraphics[height=4cm, width=4cm]{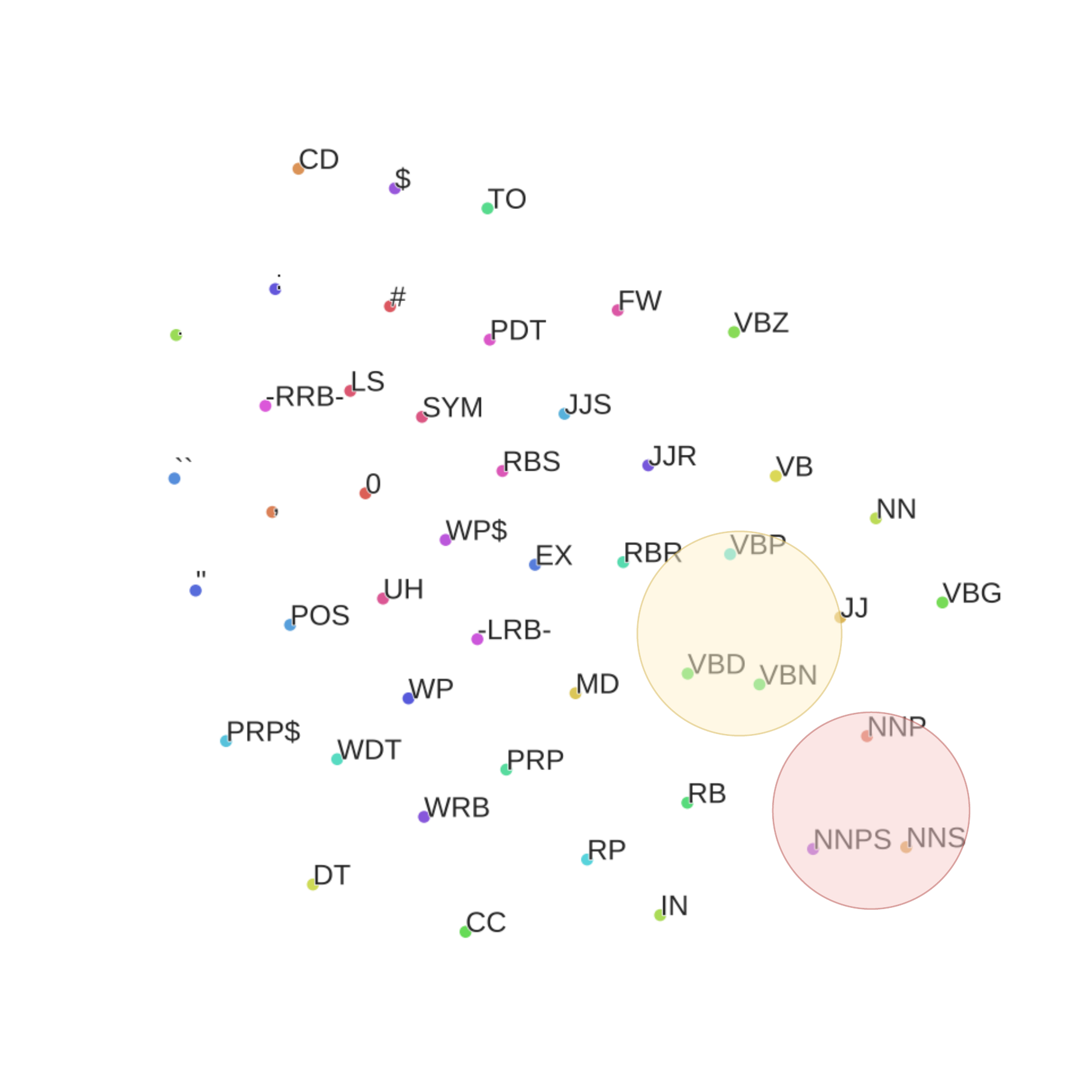}}}
\hspace{1cm}
\subfloat[15 iterations]{\fbox{\includegraphics[height=4cm, width=4cm]{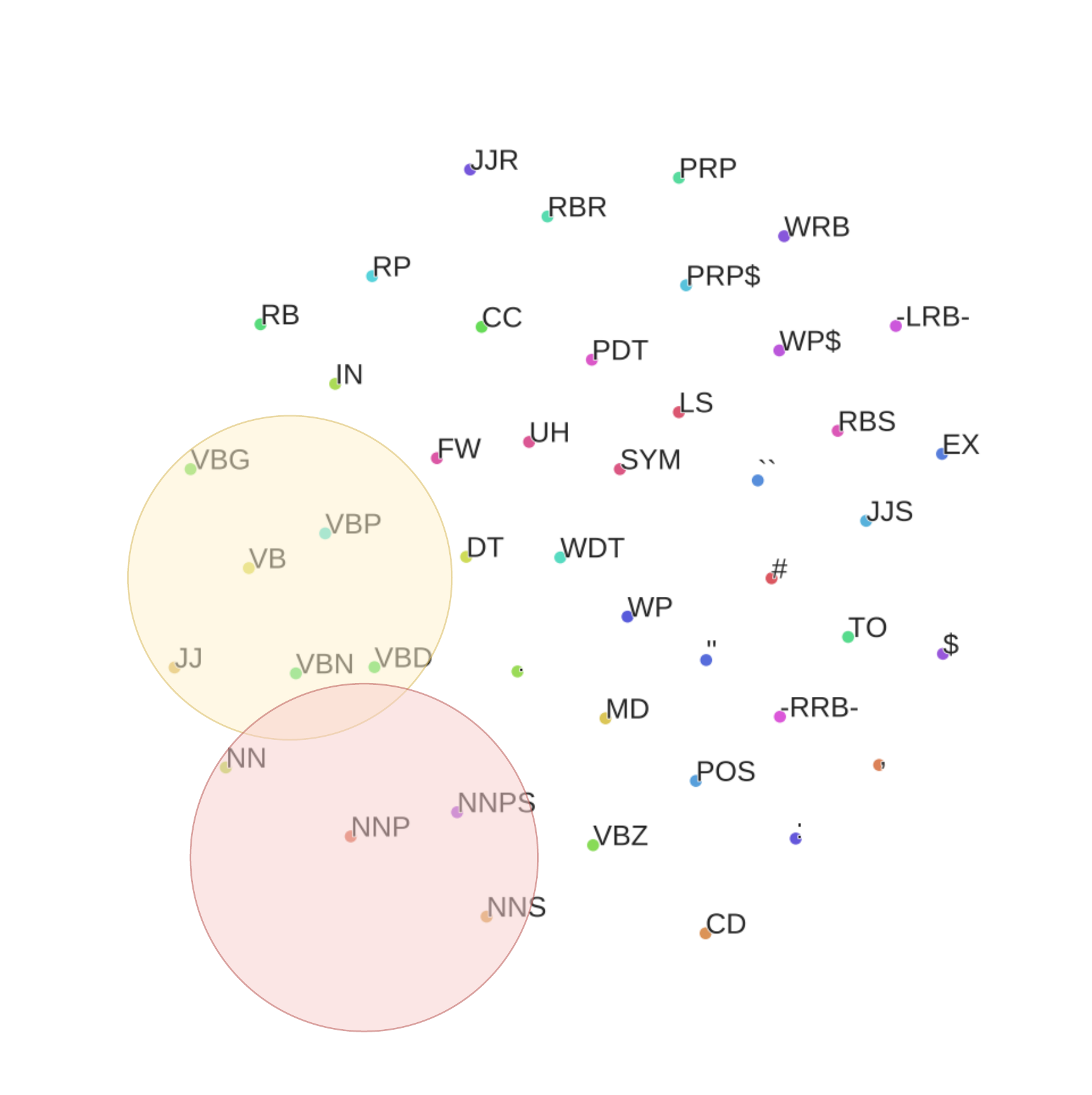}}}
\hspace{1cm}
\subfloat[38 iterations]{\fbox{\includegraphics[height=4cm, width=4cm]{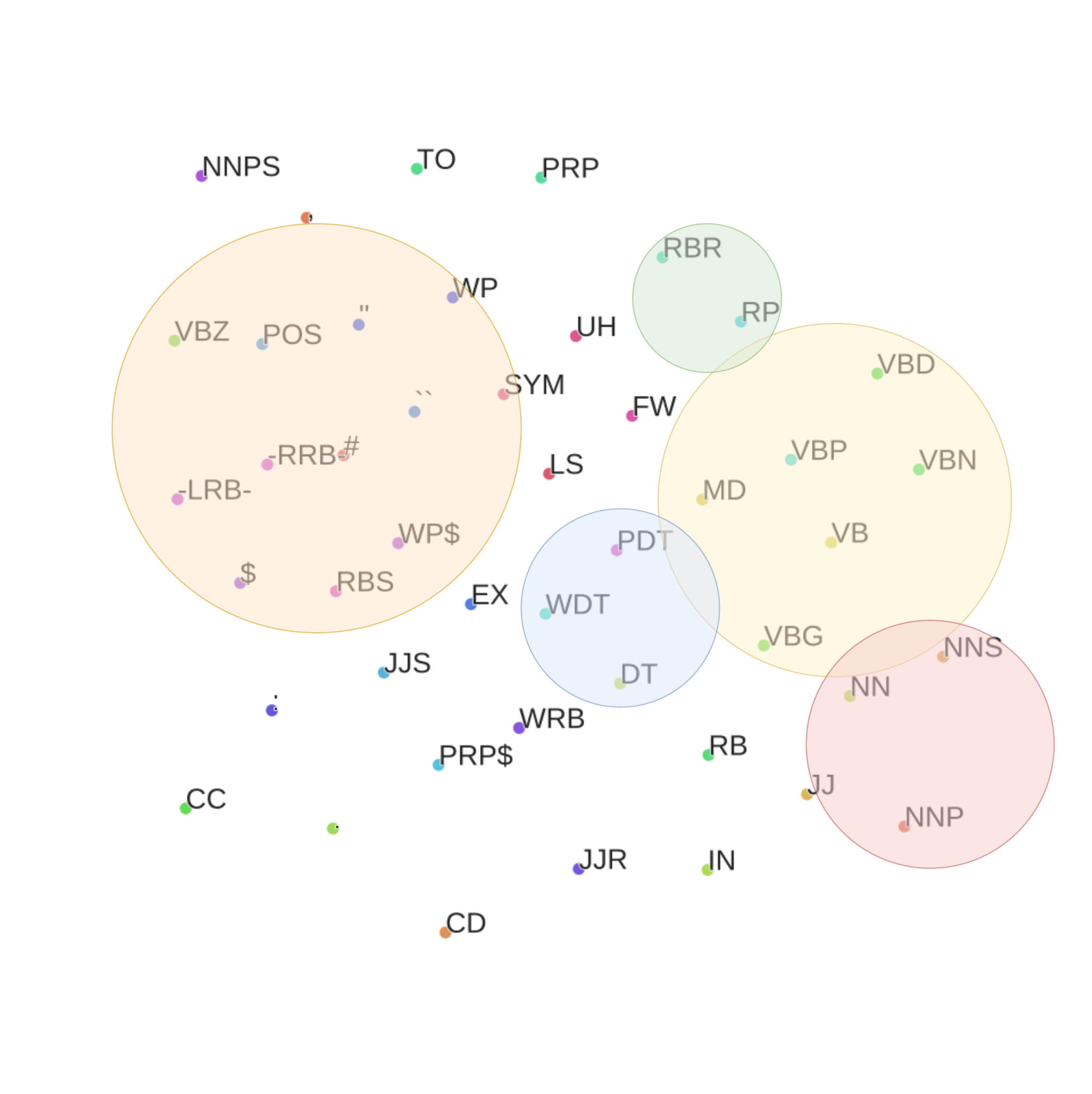}}}
\caption{\label{label_embedding} t-SNE plot of label embeddings after different numbers of training iterations.}
\end{figure*}

{\bf The effectiveness of Model Structure.}
To evaluate the effect of BiLSTM-LAN layers, we conduct ablation experiments as shown in Table~\ref{WSJ-pos-attention-lastlayer}. In BiLSTM-LAN w/o attention, we remove the attention inference sublayers from BiLSTM-LAN except for the last BiLSTM-LAN layer. This model is reminiscent to BiLSTM-softmax except that the output is based on label embeddings. It gives an accuracy slightly higher than that of LSTM-softmax, which demonstrates the advantage of label embeddings. On the other hand, it significantly underperforms BiLSTM-LAN ($p$-value$<$0.01), which shows the advantage of hierarchically-refined label distribution sequence encoding.

{\bf Model Size vs CRF.}
Table~\ref{pos-dev} also compares the effect of model sizes. We observe that: (1) As the model size increases, both BiLSTM-CRF and BiLSTM-LAN see a peak point beyond which further increase of model size does not bring better results, which is consistent with observations from prior work, demonstrating that the number of parameters is \textit{not} the decisive factor to model accuracy; and (2) the best-performing BiLSTM-LAN model size is comparable to that of the BiLSTM-CRF model size, which indicates that the model structure is more important for the better accuracy of BiLSTM-LAN.

{\bf Speed vs CRF.}
Table~\ref{pos-ccg-speed} shows a comparison of training and decoding speeds. BiLSTM-LAN processes 805 and 714 sentences per second on the WSJ and CCGBank development data, respectively, outperforming BiLSTM-CRF by 3$\%$ and 19$\%$, respectively. The larger speed improvement on CCGBank shows the benefit of lower asymptotic complexity, as discussed in Section \ref{sec:Label Attention Network}.

{\bf Training vs CRF.}
Figure~\ref{dev_accuracy_curve} presents the training curves on the WSJ development set. At the beginning, BiLSTM-LAN converges slower than BiLSTM-CRF, which is likely because BiLSTM-LAN has more complex layer structures for label embedding and attention. After around 15 training iterations, the accuracy of BiLSTM-LAN on the development sets becomes increasingly higher than BiLSTM-CRF. This demonstrates the effect of label embeddings, which allows more structured knowledge to be learned in modeling.

\subsection{Final Results}

{\bf WSJ.} 
Table~\ref{pos-result} shows the final POS tagging results on WSJ. Each experiment is repeated 5 times. 
BiLSTM-LAN gives significant accuracy improvements over both BiLSTM-CRF and BiLSTM-softmax ($p<$0.01), which is consistent with observations on development experiments. 

Table~\ref{pos-top} compares our model with top-performing methods reported in the literature. In particular, \citet{journals/corr/HuangXY15} use BiLSTM-CRF. \citet{P16-1101}, \citet{Liu2017} and \citet{yang2018design} explore character level representations on BiLSTM-CRF. \citet{P18-1030} use S-LSTM-CRF, a graph recurrent network encoder. \citet{N18-1089} demonstrate that adversarial training can improve the tagging accuracy. \citet{xin-etal-2018-learning} proposed a compositional character-to-word model combined with LSTM-CRF.
BiLSTM-LAN gives highly competitive result on WSJ without training on external data.

{\bf Universal Dependencies(UD) v2.2.}
We design a multilingual experiment to compare BiLSTM-softmax, BiLSTM-CRF (strictly following \citet{N18-1089} \footnote{\noindent Note that our results are different from Table 2
of \citet{N18-1089}, since they reported results on UD v1.2}, which is the state-of-the-art on multi-lingual POS tagging) and BiLSTM-LAN. The accuracy and training speeds are shown in Table~\ref{pos-wd}. Our model outperforms all the baselines on all the languages. The improvements are statistically significant for all the languages ($p<$0.01), suggesting that BiLSTM-LAN is generally effective across languages.

{\bf OntoNotes 5.0.}
In NER, BiLSTM-CRF is widely used, because local dependencies between neighboring labels relatively more important that POS tagging and CCG supertagging. 
BiLSTM-LAN also significantly outperforms BiLSTM-CRF by 1.17 F1-score ($p<$0.01).
Table~\ref{ner} compares BiLSTM-LAN to other published results on OntoNotes 5.0. \citet{durrett-klein-2014-joint} propose a joint model for coreference resolution, entity linking and NER. \citet{chiu-nichols-2016-named} use a BiLSTM with CNN character encoding. \citet{shen-etal-2017-deep} introduce active learning to get better performance. 
\citet{strubell-etal-2017-fast} present an iterated dilated convolutions, which is a faster alternative to BiLSTM.  
\citet{ghaddar-langlais-2018-robust} demonstrate that lexical features are actually quite useful for NER. \citet{clark-etal-2018-semi} present a cross-view training for neural sequence models. BiLSTM-LAN obtains highly competitive results compared with various of top-performance models without training on external data.

{\bf CCGBank.}
In CCG supertagging, the major challenge is a larger set of lexical tags $|L|$ and supertag constraints over long distance dependencies. As shown in Table~\ref{ccg}, BiLSTM-LAN significantly outperforms both BiLSTM-softmax and BiLSTM-CRF ($p<$0.01), showing the advantage of LAN. \citet{P15-2041} and \citet{a11} explore BiRNN-softmax and BiLSTM-softmax, respectively. \citet{P16-2038} present a multi-task learning architecture with BiRNN to improve the performance. \citet{N16-1026} train BiLSTM-softmax using tri-training. \citet{N16-1027} combine a LSTM language model and BiLSTM over the supertags.
\citet{tu-gimpel-2019-benchmarking} introduce the inference network \cite{tu-18} in CCG supertagging to speed up the training and decoding for BiLSTM-CRF.
Compared with these methods, BiLSTM-LAN obtains new state-of-the-art results on CCGBank, matching the tri-training performance of \citet{N16-1026}, without training on external data.

\begin{table}[t!]
\small
\begin{center}
\begin{tabular}{lc}
\hline \bf Model & \bf F1 \\
\hline
\citet{durrett-klein-2014-joint} & 84.04 \\
\citet{chiu-nichols-2016-named} & 86.28 \\
\citet{shen-etal-2017-deep} & 86.52 \\
\citet{strubell-etal-2017-fast} & 86.84 \\
\citet{ghaddar-langlais-2018-robust} & 87.95 \\
\citet{clark-etal-2018-semi}$^{*}$ & \bf{88.81} \\
\hline
BiLSTM-softmax \cite{strubell-etal-2017-fast} & 83.76 \\
BiLSTM-CRF \cite{strubell-etal-2017-fast}& 86.99 \\
BiLSTM-LAN & 88.16 \\
\hline

\end{tabular}
\end{center}
\caption{\label{ner} F1-scores on the OntoNotes 5.0 test set. * denotes semi-supervised and multi-task learning.
}
\end{table}

\begin{table}[t!]
\small
\begin{center}
\begin{tabular}{lc}
\hline \bf Model & \bf Accuracy (\%) \\
\hline
\citet{P15-2041}  & 93.0 \\
\citet{P16-2038}  & 93.3 \\
\citet{a11} & 94.2 \\
\citet{N16-1026} & 94.3  \\
\citet{N16-1026} & \ \! 94.7$^*$  \\
\citet{N16-1027} & 94.5 \\
\citet{tu-gimpel-2019-benchmarking} & 94.4 \\
\hline
BiLSTM-softmax & 94.1 \\
BiLSTM-CRF & 94.1 \\
BiLSTM-LAN & \bf{94.7} \\
\hline

\end{tabular}
\end{center}
\caption{\label{ccg} Supertagging accuracy on CCGbank test set. * indicates that further gains follow from semi-supervised tri-training (improving the accuracy from 94.3\% to 94.7\%).
}
\end{table}
\begin{table*}[t!]
\small
\begin{center}
\begin{tabular}{c|c|c|c|c}
\hline \bf Sentence & \bf Gold Standard & \bf LSTM-softmax &\bf LSTM-CRF & \bf LSTM-LAN \\
\hline
it & NP & NP & NP & NP\\
settled & (S[dcl]\textbackslash NP)$/$PP & (S[dcl]\textbackslash NP)$/$PP & (S[dcl]\textbackslash NP)$/$PP & (S[dcl]\textbackslash NP)$/$PP \\
with & ((S\textbackslash NP)\textbackslash(S\textbackslash NP))$/$NP & \cellcolor{yellow}　PP$/$NP & \cellcolor{yellow}　PP$/$NP & ((S\textbackslash NP)\textbackslash(S\textbackslash NP))$/$NP\\
a & NP[nb]$/$N & NP[nb]$/$N & NP[nb]$/$N & NP[nb]$/$N \\
loss & N & N & N & N \\
of & (NP\textbackslash NP)$/$NP & (NP\textbackslash NP)$/$NP & (NP\textbackslash NP)$/$NP & (NP\textbackslash NP)$/$NP \\
4.95 & N $/$ N & N $/$ N & N $/$ N & N $/$ N \\
cents & N & N & N & N \\
at & PP$/$NP & \cellcolor{yellow} (NP\textbackslash NP)$/$NP & PP$/$NP & PP$/$NP \\
$\$$ & N $/$ N[num] & N $/$ N[num] & N $/$ N[num] & N $/$ N[num] \\
1.3210 & N[num] & N[num] & N[num] & N[num]\\
a & (NP\textbackslash NP) $/$ N & (NP\textbackslash NP) $/$ N & (NP\textbackslash NP) $/$ N & (NP\textbackslash NP) $/$ N \\
pound & N & N & N & N \\
.&.&.&.&.\\
\hline
\end{tabular}
\end{center}
\caption{\label{ccg_case_study} CCG case analysis. The error are in yellow.
}
\end{table*}

\section{Discussion}
\begin{figure}[t]
\setlength{\belowcaptionskip}{-6pt}
\centering
\includegraphics[width=8.5cm]{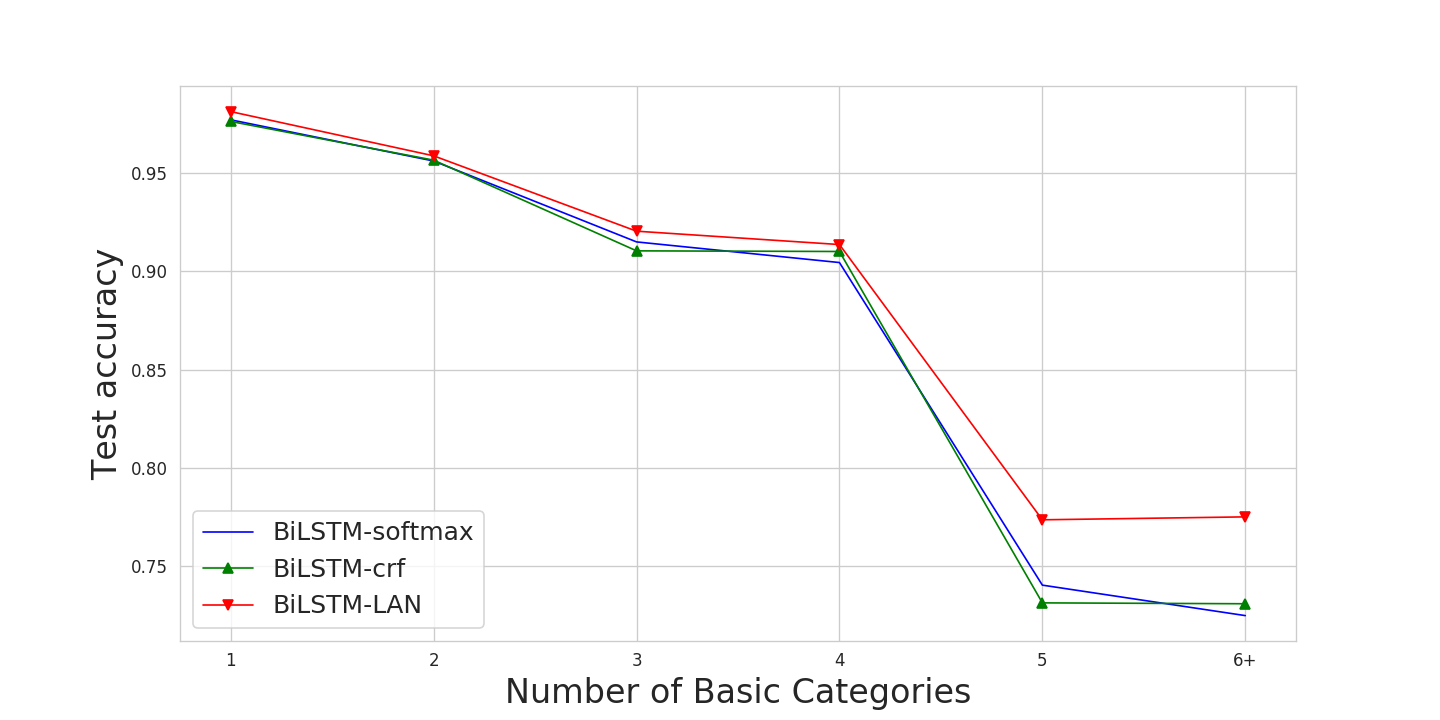}\caption{\label{ccg_basic_cate} Accuracy against supercategory complexity.}
\end{figure}

{\bf Visualization.}
A salient advantage of LAN is more interpretable models. We visualize the label embeddings as well as label attention weights for POS tagging.
We use t-SNE to visualize the 45 different English POS tags (WSJ) on a 2D map after 5, 15, 38 training iteration, respectively. Each dot represents a label embedding.
As can be seen in Figure~\ref{label_embedding}, label embeddings are increasingly more meaningful during training. Initially, the vectors sit in random locations in the space. After 5 iterations, small clusters emerge, such as ``NNP'' and ``NNPS'', ``VBD'' and ``VBN'', ``JJS'' and ``JJR'' etc. The clusters grow absorbing more related tags after more training iterations. 
After 38 training iterations, most similar POS tags are grouped together, such as ``VB'', ``VBD'', ``VBN'', ``VBG'' and ``VBP''. More attention visualization are shown in Appendix B.

{\bf Supercategory Complexity.}
We also measure the complexity of supercategories by the number of basic categories that they contain. According to this definition, ``S'', ``S$/$NP'' and ``(S \textbackslash NP)$/$NP'' have complexities of 1, 2 and 3, respectively. Figure~\ref{ccg_basic_cate} shows the accuracy of BiLSTM-softmax, BiLSTM-CRF and BiLSTM-LAN against the supertag complexity. As the complexity increases, the performance of all the models decrease, which conforms to intuition. BiLSTM-CRF does not show obvious advantages over BiLSTM-softmax on complex categories. In contrast, BiLSTM-LAN outperforms both models on complex categories, demonstrating its advantage in capturing more sophisticated label dependencies.

{\bf Case Study.}
Some predictions of BiLSTM-softmax, BiLSTM-CRF and BiLSTM-LAN are shown in Table~\ref{ccg_case_study}. The sentence contains two prepositional phrases ``with ...'' and ``at ...'', thus exemplifies the PP-attachment problem, one of the hardest sub-problems in CCG supertagging. As can be seen, BiLSTM-softmax fails to learn the long-range relation between ``settled'' and ``at''. BiLSTM-CRF strictly follows the hard constraint between neighbor categories thanks to Markov label transition. However, it predicts ``with" incorrectly as ``PP$/$NP" with the former supertag ending with ``$/$PP". In contrast, BiLSTM-LAN can capture potential long-term dependency and better determine the supertags based on global label information. In this case, our model can effectively represent a full label search space without making Markov assumptions.


\section{Conclusion}

We investigate a hierarchically-refined label attention network (LAN) for sequence labeling, which leverages label embeddings and captures potential long-range label dependencies by deep attentional encoding of label distribution sequences. Both in theory and empirical results prove that BiLSTM-LAN effective solve label bias issue. Results on POS tagging, NER and CCG supertagging show that BiLSTM-LAN outperforms BiLSTM-CRF and BiLSTM-softmax.

\section*{Acknowledgments}

We thank Zhiyang Teng and Junchi Zhang for insightful discussions. We thank Chenhua Chen for proofreading the paper. We also thank all anonymous reviewers for their constructive comments. 
This work is supported by National Science Foundation of China (Grant No. 61976180).
The corresponding author is Yue Zhang.
\bibliographystyle{acl_natbib}
\bibliography{emnlp-ijcnlp-2019}

\clearpage
\onecolumn
\begin{appendices}
\section{Hyper-Parameters}
\begin{table}[h]
\small
\begin{center}
\begin{tabular}{l|r}
\hline \bf Hyper-parameter & \bf Value \\
character embeddings & 30 \\
character-level hidden size & 50 \\
word-level hidden size (English) & 400\\ 
word-level hidden size (other languages) & 200\\ 
encoder layers (POS) & 3 \\
encoder layers (NER) & 4 \\
encoder layers (CCG) & 5 \\
number of attention heads & 5 \\
drop out & 0.5 \\
batch size & 10 \\
optimizer & SGD \\
momentum & 0.9 \\
L2 regularization & $1e^{-8}$ \\
learning rate & 0.01 \\
decay rate (POS) & 0.035 \\
decay rate (NER) & 0.04 \\
decay rate (CCG) & 0.05 \\
gradient clipping & 5.0 \\
\hline

\end{tabular}
\end{center}
\caption{\label{hyper-parameters} Hyper-parameters.
}
\end{table}

Tabel~\ref{hyper-parameters} shows the hyper-parameters of BiLSTM-LAN used for all the experiments in the papers.

\section{Attention Visualization}
\begin{figure*}[!h]
\centering
\includegraphics[width=16cm]{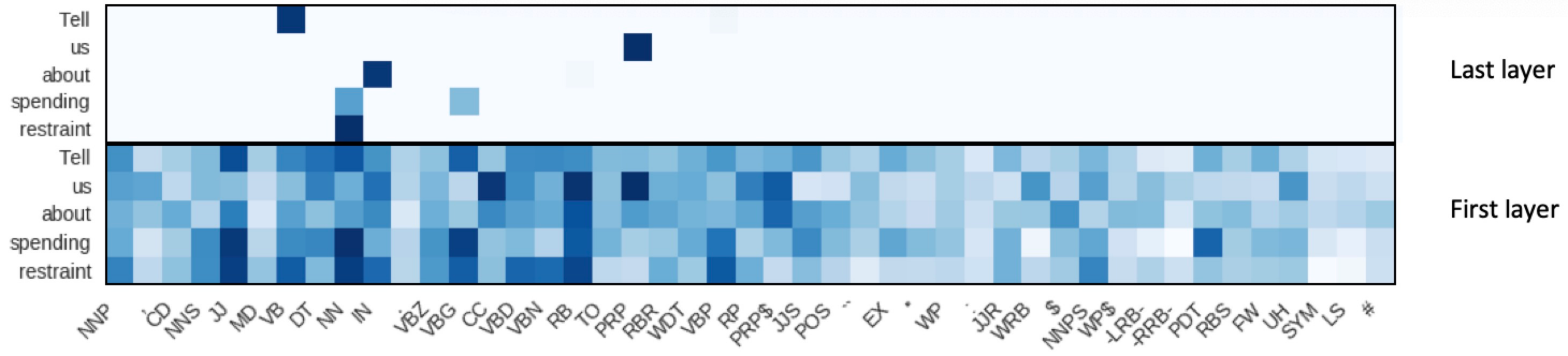}
\caption{\label{attentionWeight} Attention visualizations for first and last attention layer, respectively. The color depth expresses the word's preference degree for each label in attention vector.}
\end{figure*}
Here we continue the analysis of visualization from Section 6, we also visualize the label attention weights of BiLSTM-LAN, given the sentence in ``Tell (VB) us (PRP) about (IN) spending (NN) restraint (NN)". As shown in Figure~\ref{attentionWeight}, the first layer contains initial label distributions according to unigram information, in which much ambiguity remains. For example, likely POS tags for the word ``{\textit{spending}}''  include ``JJ'', ``NN", ``VBG'', ``TO'' and ``RBS'', while the attention layer assigns some probabilities to almost every label. In the second layer, the label distribution of every word becomes sharp and concentrated on the most probable tags in the sentential context. Here the word ``spending'' bares only ``NN'' and ``VBG'' labels, with relatively more probability (0.114) being assigned to ``NN''.
\end{appendices}

\end{document}